# Decoding CNN based Object Classifier Using Visualization

Abhishek Mukhopadhyay[†]
  Indian Institute of Science, Bengaluru, 560012, India, abhishekmukh@iisc.ac.in

Imon Mukherjee
  Indian Institute of Information Technology, Kalyani, 741235, India, imon@iiitkalyani.ac.in

Pradipta Biswas
  Indian Institute of Science, Bengaluru, 560012, India, pradipta@iisc.ac.in

[†] Abhishek Mukhopadhyay affiliated to both Indian Institute of Science and Indian Institute of Information Technology Kalyani.

## ABSTRACT

This paper investigates how working of Convolutional Neural Network (CNN) can be explained through visualization in the context of machine perception of autonomous vehicles. We visualize what type of features are extracted in different convolution layers of CNN that helps to understand how CNN gradually increases spatial information in every layer. Thus, it concentrates on region of interests in every transformation. Visualizing heat map of activation helps us to understand how CNN classifies and localizes different objects in image. This study also helps us to reason behind low accuracy of a model helps to increase trust on object detection module.

## CCS CONCEPTS

•Computing methodologies~Artificial intelligence~Computer vision~Computer vision problems~Object detection •Human-centered computing~Visualization~Empirical studies in visualization

## 1 Introduction

In recent time, significant progress has been made in Autonomous Driving Assistance System (ADAS), which are capable of sensing and reacting to its immediate environment. The task of environment sensing is known as perception and consists of several subtasks such as semantic segmentation, object detection and classification. Object detection allows ADAS to recognize traffic signs, traffic lights, cars, lanes, pedestrians and so on. Progress in Convolutional Neural Network (CNN) based object detection methods (YOLOv3, Multi-Box Single Shot Detector) help to detect objects in real time [1, 3]. This paper focuses on explaining how intermediate layers of a CNN works and which part of the image has highest influence in final prediction as part of developing object detection model for real time deployment.

Zeiler and Fergus [5] used deconvolution technique to show what type of pattern in an input image activates specific set of neurons which helped them to change architecture of CNN to perform state of the art performance on ImageNet 2012 validation dataset. Simonyan [9] demonstrated how to obtain saliency maps of convolution layer (ConvNet) classification models using numerical optimization of the input image. They also demonstrated how to obtain saliency maps of ConvNet classification models by projecting back from the fully connected layers of the network for a given input image. Girshick [4] used visualization to check which part of proposed region are responsible for strong activations at higher layers in their object detection model. Bojarski [2] introduced 'VisualBackProp' technique for visualizing which part of image attribute most to the prediction of CNN. They used it as debugging tool for steering self-driving cars in real time. Yosinski [10] developed two tools for visualizing output of CNNs. In the first tool, they visualized activation produced in each convolution layer for any input images or videos. In the second tool, they introduced several regularizations to produce better interpretable visualization compare to first tool. Rieke [7] used visualization to confirm whether trained model focused on the object in image in prediction time. They trained a 3D CNN using MRI scanned images for detecting Alzheimer's disease. Later they used four different gradient-based and occlusion-based techniques to visualize which part of image activating CNN most. Despite the encouraging progress in visualization techniques, there is a scope for integrating these techniques with real time applications to interact with CNN models. In this paper we have used two visualization techniques to understand working of CNN to optimize their architecture to predict road objects in real time for autonomous vehicle.

We have visualized intermediate ConvNet outputs to interpret how successive convolution layers' transform input images. We have also used Gradient-weighted Class Activation Mapping (Grad-CAM) technique [8] to understand which part of the object is responsible for localizing the object's position in an image. These steps can be used to increase trust on object detection in visual spectrum using CNN and to develop efficient object tracking algorithms.

## 2  Proposed Approach

The proposed visualization system works in twofold: (I) Visualizing intermediate ConvNet outputs (intermediate activations) is useful to understand how successive convnet layers transform their input. It also gives us idea of what type of features are extracted by different filters of different layers of CNN model from input images. (II) Visualizing heatmap using Grad-CAM techniques [8] to see which part of image led final convolution layer to predict object in the image. Our aim is to build object detection model for both Indian road environment and automated taxing of airplane. In order to do so, we mixed both type of objects found in road and airfield i.e. car, motorbike, bi-cycle, and airplane for training and testing. We used state-of-the-art VGG16 model which was pre-trained on ImageNet dataset. We prepared dataset by mixing Indian and western road images downloaded from Kaggle and Google Images. We trained the model on total 2823 images parted in training and validation dataset (70:20) for 50 epochs with a batch size of 32. In order to understand how CNN model can classify the input image, we need to understand how our model sees the input image by looking at the output of its intermediate layers. We visualized activations in $\left(\frac{n}{4}\right)th$ convolutional layer, $\left(\frac{n}{2}\right)th$ convolutional layer, $\left(\frac{3n}{4}\right)th$ convolutional layer, and $n-th$ convolutional layer of the trained model. We also visualized heatmap of class activation over test images.

We visualized three different group of objects, i.e. road objects in western road (car, motorbike), road objects in Indian road (car, motorbike, cycle), and airplane (real and synthetic images). We found that first few convolution layers of the model extracted basic features (edges, contours) of the object and retained maximum information from the input image (Figure 1(b)). As we go deeper in the model, activations became less visually interpretable (Figure 1(c) – 1(e)). Model started to extract abstract features (e.g., patch-based features like skeleton of a car). In the deeper level of the network resolution of feature map starts decreasing but spatial information increases.

If we observe all four-feature map outputs (Figure 1(c) – 1(e)), it is evident that in each transformation model eliminated background or irrelevant information and refined useful information related to class of objects. We also visualized every channel of activations on the test images. We found that in first few layers (Figure 2(a)) almost all feature maps were activated but with deeper layers, number of dead feature maps (blue colored boxes in Figure 2) started increasing (Figure 2(b)). This might happen due to lack of information presented in the input images.

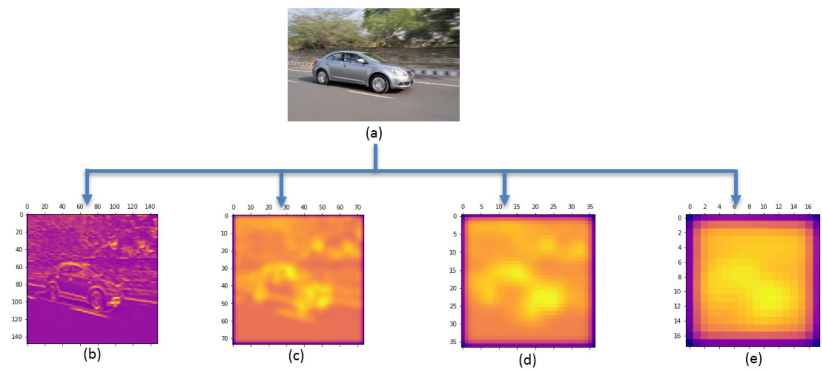

**Figure 1:** (a) Input test image on Indian road; (b) 28th channel of the activation of 3rd convolution layer; (c) 28th channel of the activation of 7th convolution layer; (d) 28th channel of the activation of the 10th convolution layer; (e) 510th channel of the activation of the 13th convolution layer. This figure is best viewed in electronic form.

We also visualized heatmap of class activation to understand which part of the object were responsible to let the model to classify properly. In this context, class activation map tells us which part of an image correspond to a class of object. In Figure 3 we showed heatmap of car and motorbike. We found that bonnet of car, tire and engine of motorbike were strongly activated irrespective of their color in the images (where red color indicate highest gradient score and cyan indicate least gradient score). It helped us to understand whether model predicted proper class of object or not. It also gave insight about the performance of the model. It may happen that model give higher accuracy during training but fail miserably in testing time. It happens due to overfitting of the model during training phase. Grad-CAM based heatmap help us to understand whether model can able to locate the object in the image or not. Earlier work found existing CNN based object detection models have limited success in novel traffic participants like for traffic participants in Indian road scenario [6]. Our visualization work aims to develop new object detection model for Indian road context to address those limitations.

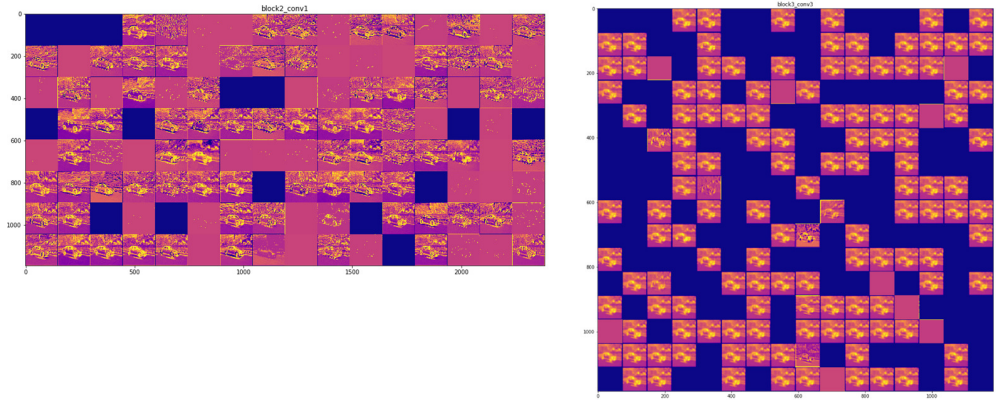

**Figure 2:** All channels of convolutional layer activation. (a) activation of 3$^{rd}$ convolution layer; (b) activation of 7$^{th}$ convolution layer. Figures are best viewed in electronic form and numbered in clockwise manner.

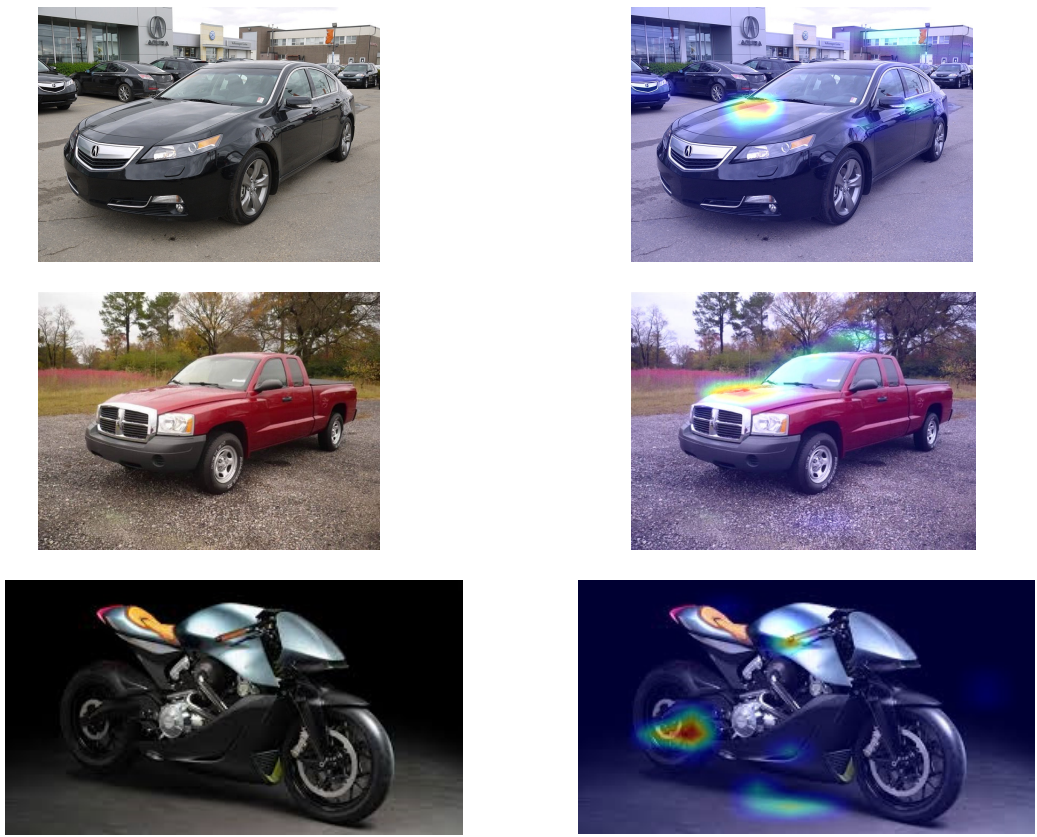

**Figure 3:** All channels of convolutional layer activation. (a) activation of 3$^{rd}$ convolution layer; (b) activation of 7$^{th}$ convolution layer. Figures are best viewed in electronic form and numbered in clockwise manner.

# 3 Conclusion

The proposed visualization system works in t this paper, we have applied different visualization methods to understand classifier decisions of a CNN. This work is part of building a lightweight object detection model that can interpret outside situations in real time in autonomous vehicle. In order to do so, we visualized how CNN interpret input images in different layers and how does it eliminate unnecessary information from the image in consequent layers to focus on region of interest in the image. Heatmap based visualization help us to investigate which part of image contributes in case of false positive results. This will help us to trust on CNN based object detection model where security is utmost priority for driver and co-passengers in the vehicle.